\documentclass[letterpaper, 12 pt, conference]{ieeeconf}

\IEEEoverridecommandlockouts                              

\usepackage{lipsum}
\usepackage{graphicx}                       
\usepackage{stfloats}
\usepackage[tight,footnotesize]{subfigure}  

\usepackage{amssymb,amsmath}
\usepackage{textcomp}
\usepackage{mdwmath}
\usepackage{mdwtab}
\usepackage{eqparbox}
\usepackage{rotating}                       
\usepackage{array}                          
\usepackage{listings}                       
\usepackage[ruled,vlined,noresetcount]{algorithm2e}
\usepackage{listings}
\usepackage[noend]{algpseudocode}

\usepackage[colorlinks,
            linkcolor=blue,
            anchorcolor=blue,
            citecolor=blue,
            bookmarks=false
            ]{hyperref}
\usepackage{cite}

\usepackage{graphicx}
\usepackage{float}
\usepackage{subfigure}
\usepackage{amsmath}
\usepackage{algpseudocode}
\usepackage{stfloats}
\usepackage{multirow} 
\usepackage{xcolor}
\usepackage[ruled,vlined]{algorithm2e}
\usepackage{amssymb}
\usepackage{booktabs}

\title{\LARGE \bf A Light-Weight LiDAR-Inertial SLAM System with Loop Closing}

\author{Kangcheng Liu and Huosen Ou
\thanks{ Kangcheng Liu is the corresponding author.}
}

\begin{document}

\maketitle
\thispagestyle{empty}
\pagestyle{empty}

\begin{abstract}
In this work, we propose a lightweight integrated LiDAR-Inertial SLAM system with high efficiency and a great loop closure capacity. We found that the current State-of-the-art LiDAR-Inertial SLAM system has poor performance in loop closure. The LiDAR-Inertial SLAM system often fails with the large drifting and suffers from limited efficiency when faced with large-scale circumstances. In this work, firstly, to improve the speed of the whole LiDAR-Inertial SLAM system, we have proposed a new data structure of the sparse voxel-hashing to enhance the efficiency of the LiDAR-Inertial SLAM system. Secondly, to improve the point cloud-based localization performance, we have integrated the loop closure algorithms to improve the localization performance. Extensive experiments on the real-scene large-scale complicated circumstances demonstrate the great effectiveness and robustness of the proposed LiDAR-Inertial SLAM system.

\end{abstract}

\section{Introduction}
The robotics 3D vision technologies have emerged as the highly effective ways to perceive and interact with the real physical world \cite{liu2020fg}. The 3D vision has enabled us to detect the interested objects and segment the objects we are interested in at the pixel level \cite{liu2022fg, liu2022ws3d, liu2022integrated2}.  The Light Detection And Ranging Sensor (LiDAR) is the most popular 3D sensor. The fast 3D reconstruction and mapping is of great significance to many applications, such as the robotics navigation, the robotics surveillance applications, and the robotics autonomous mapping of large-scale areas for the applications such as search and rescue in very complicated circumstances. The LiDAR Inertial fusion-based approaches have great potentials in power grid inspections and building inspections \cite{liu2017avoiding, liu2022industrial}, industrial infrastructural inspections \cite{liu2019deep}. The most popular and typical LiDAR-SLAM system includes LOAM \cite{zhang2014loam, liu2022dlc, liu2022robust2}, LEGO-LOAM \cite{shan2018lego, liu2022robust, liu2022datasets}, and LIO-SAM \cite{shan2020lio}. However, most LiDAR-Inertial SLAM algorithms take approximately 100 ms to process a single LiDAR scan \cite{liu2022enhanced}, with a speed of around 10Hz. With the development of recent LiDAR mapping technologies, the point clouds capturing frequency by the customized LiDAR such as Livox solid-state LiDAR can reach more than 100 Hz recently. However, the advancement of current LiDAR-Inertial SLAM system can not catch up with the updating of the consumer-usable LiDAR hardware. In this work, one important aspect of our focuses is improving the efficiency of the LiDAR-Inertial SLAM system. For an actual integrated autonomous navigation system, the LiDAR-Inertial SLAM algorithms for localization and mapping is not the only components of the system. The computational resources are shared by various of autonomous robotics components such as task planning, the motion planning, the object detection and localization \cite{liu2021fg}, the visual tracking, and advanced control techniques. All the algorithms are jointly running on a GPU-based micro-computer with limited resource such as the NVIDIA Jetson TX2 and Jetson Xavier. \\
\begin{figure}[tbp!]
\centering
\includegraphics[scale=0.21]{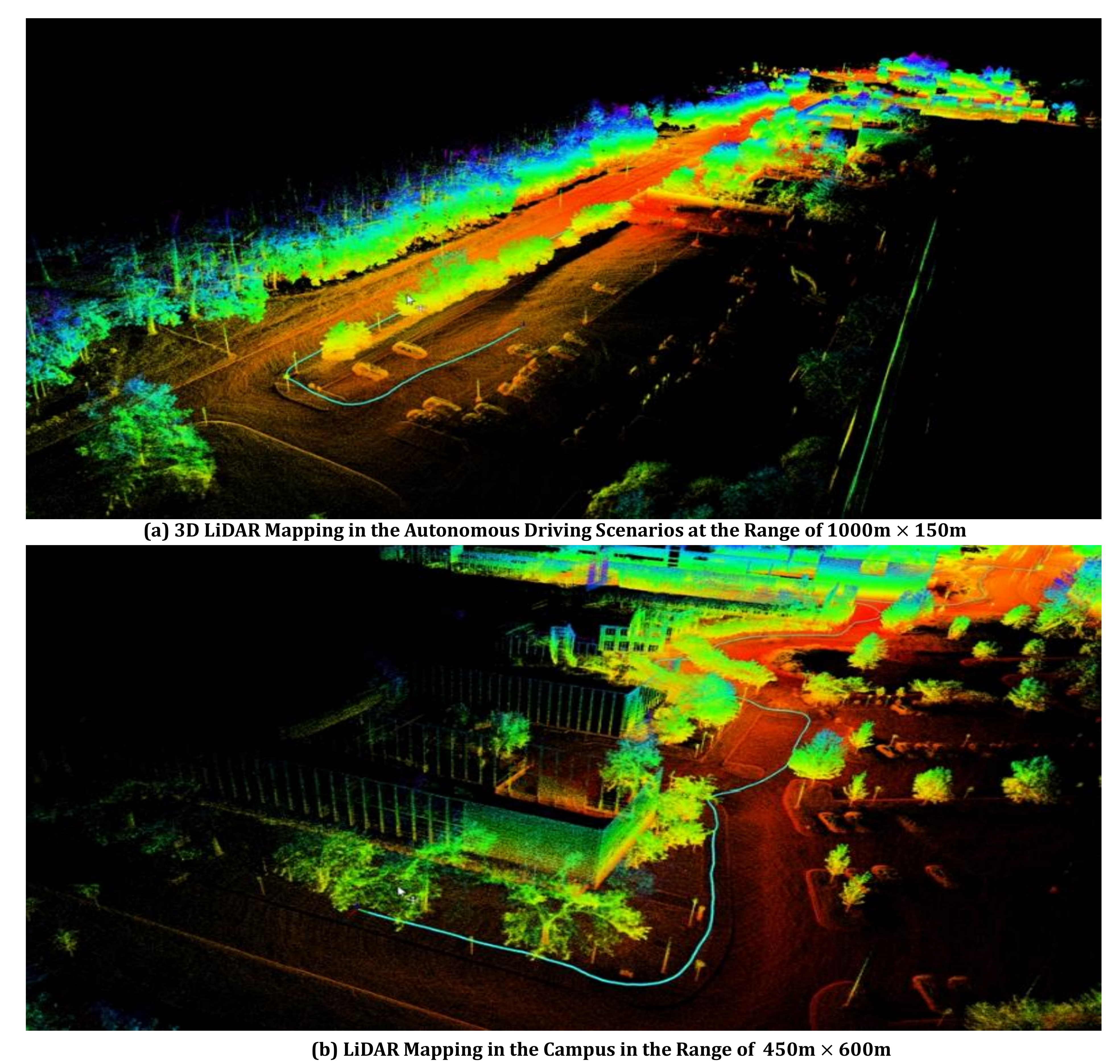}
\caption{The real-site testing of our proposed LiDAR-Inertial SLAM system at diverse large-scale real-site test circumstances. The Subfigure (a) shows the mapping results of an autonomous driving scenarios at the range of $1000m \times 150m$. The Subfigure (b) shows the mapping results at the range of $450m \times 600m$.}
\vspace{-5.9mm}
\label{fig_overall}
\end{figure}  
Therefore, the LiDAR-Inertial odometry system should be as light-weight as possible to ensure the high-level of efficiency and robustness of the whole robotics system. From our experiments, the efficiency of the LiDAR-Inertial odometry system mainly depends on the point clouds searching and the backend odometry, which are summarized in details in the related work. In this work, we have proposed a novel data structure that can handle well with the the point clouds searching compared with the previous KD-Tree \cite{chen2019fast} and VDB-based \cite{museth2013vdb} methods. \\
Also, the loop closure recognition is an important problem that need to be tackled when doing LiDAR-based mapping. The learning-based approaches have been explored extensively in image processing \cite{yuzhi2020legacy, liu2022integrated, liu2022semi}, but remain rarely explored in the fields of robotics 3D localization and mapping. In the real-scene complicated circumstances, it is very difficult for the autonomous robots to accurately find the loop closures. For examples, in the complex environments as shown in Fig. \ref{fig_overall}, the circumstances are very large with vairous of background objects, and sometimes the loop closures can not be easily detected. Due to the strong capacity of feature extraction of deep learning-based methods, in this work, we have proposed a novel network framework, term \textit{FG-LC-Net}, to tackle the problem of loop closure in LiDAR-Inertial SLAM.
In summary, to improve the robustness, accuracy and efficiency of the whole LiDAR SLAM system and tackle the challenges mentioned above, in this work, we have proposed a light-weight LiDAR-Inertial SLAM system with high efficiency and loop closure detection capacity. As shown in Fig. \ref{fig_overall}, it has realized accurate and efficient mapping of various real-scene complicated indoor, outdoor, and the underground tunnel subterranean circumstances. In summary, we have the following three prominent contributions:
\begin{enumerate}
    \item  Firstly, we have proposed an efficient data structure termed \textit{S-Voxel} to improve the speed of the nearest neighbour query. Extensive experiments demonstrate the proposed \textit{S-Voxel} is highly efficient and improve the efficiency of the whole SLAM system.
    \item  Secondly, we have proposed a light-weight network framework termed \textit{FG-LC-Net} \cite{liu2020fg}. It has been demonstrated by extensive experiments that our proposed loop closure detection methods have great advantages in improving the overall accuracy of the LiDAR SLAM system in complicated circumstances. 
    
    \item  Finally, compared with many previous methods which are merely demonstrated by the simulations and simple application cases, we have done extensive real-site experiments in the indoor, outdoor, and the underground tunnel subterranean environments to demonstrate the effectiveness of our proposed method.
\end{enumerate}

Our SLAM system has been tested under various circumstances including indoor, outdoor, and tunnel environments.

\vspace{-1.1mm}         
\section{Related Work} 
According to our experiments, the efficiency of the LiDAR-Inertial odometry system mainly depends on the front-end point clouds searching and the backend odometry. 
We summarize the three aspects that limit the efficiency of the LiDAR SLAM system. Also, we have summarized some typical related work in the loop closure detection of the LiDAR-Inertial SLAM system.

\subsection{The Nearest Neighbour Calculation}
 The basic problem in point clouds registration is to calculate the nearest neighbour of the given point to the historical point clouds. It often relies on some nearest neighbor data structure. The data structure can be roughly divided into the tree-like and the voxel-like data structure.  In a broad sense, the high-dimensional nearest neighbor problem is a more complex problem, but the nearest neighbor in LiDAR-Inertial Odometry (LIO) is a low-dimensional and incremental problem. Therefore, static data structures like R* trees, B* trees, etc. are not very suitable for the LIO. In Fast-LIO2 \cite{xu2021fast}, it is proposed to use incremental kdtree to deal with nearest neighbors, and we believe that incremental voxels are more suitable for LIO systems.
\subsection{The Calculation Manner of the Point Clouds Residual} 

In autonomous driving, there is a general preference not to directly utilize point-to-point residuals directly, but to use point-to-line or point-to-surface residuals. Although the point-to-point residual form is simple, the radar point cloud is more sparse than the RGB-D point cloud, and the same point may not be hit during the vehicle movement process, and the point cloud is often down-sampled before processing, so it is not very suitable for use in autonomous driving. The LOAM series will calculate point cloud features. In practice, the time spent on feature extraction is not negligible, even the main calculation part.

\subsection{The Selection of the State Estimation Algorithms}
In the commonly adopted LIO or visual inertial odometry, the schemes between single-frame EKF and batch optimization are generally used. Schemes between single-frame EKF and batch optimization are generally used in LIO and VIO, such as the sliding window filter \cite{huang2011observability}, Iterative-EKF \cite{xu2021fast}, MSCKF \cite{mourikis2007multi}, Sliding Window Filter \, and so on. Among them, Iterative-EKF is the simplest and most effective category, which not only ensures the accuracy through iteration, but also does not need to calculate a bunch of Jacobian matrices like the pre-integration system. Recently, many LIO integrated with vision systems have been proposed \cite{lin2021r3live}. In these systems, the Iterative-EKF \cite{xu2021fast} is the most effective one for state estimation. This is because that the Iterative-EKF \cite{shan2020lio} is the simplest and most effective category among them, which not only ensures the accuracy through iteration, but also does not need to calculate a bunch of Jacobian matrices like the pre-integration system.  

\subsection{The Loop Closure Detection of the LiDAR SLAM System}
The loop closure detection is also of great significance to large-scale robotics localization in very complicated circumstances. 
The loop closure detection methods can be divided into vision-based and LiDAR-based approaches. A typical vision-based method is the bag of words model \cite{galvez2012bags}, which measures the smallest distance of visual words to the trained vision vocabulary. Superglue \cite{sarlin2020superglue}-based learned feature has also been introduced, but it suffers from limited domain adaptation capacity, such as the transition between indoor and outdoor scenarios. The LPD-Net \cite{liu2019lpd} and OverlapNet  \cite{chen2021overlapnet} are recently proposed point cloud-based networks for the place recognition, and the scan-context \cite{kim2018scan} based approach has also been proposed to perform loop closure.  However, their real-world performance in complicated circumstances is limited according to our experiments. 

\begin{figure}[t]
\centering
\includegraphics[scale=0.25]{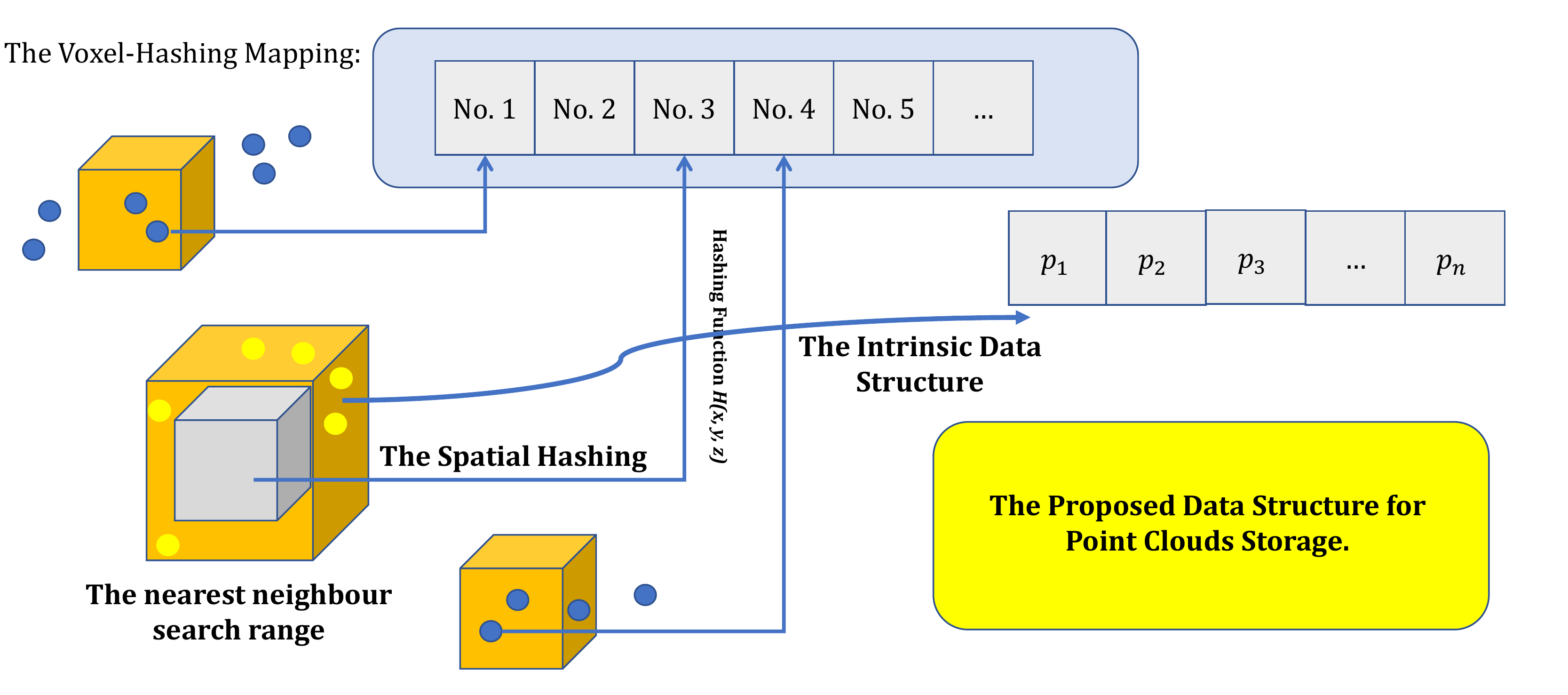}
\caption{The proposed $\textit{S-Voxel}$ data structure to improve the nearest neighbor query efficiency.}
\vspace{-3.6mm}
\label{fig_svoxel}
\end{figure}

\begin{figure}[t]
\centering
\includegraphics[scale=0.259]{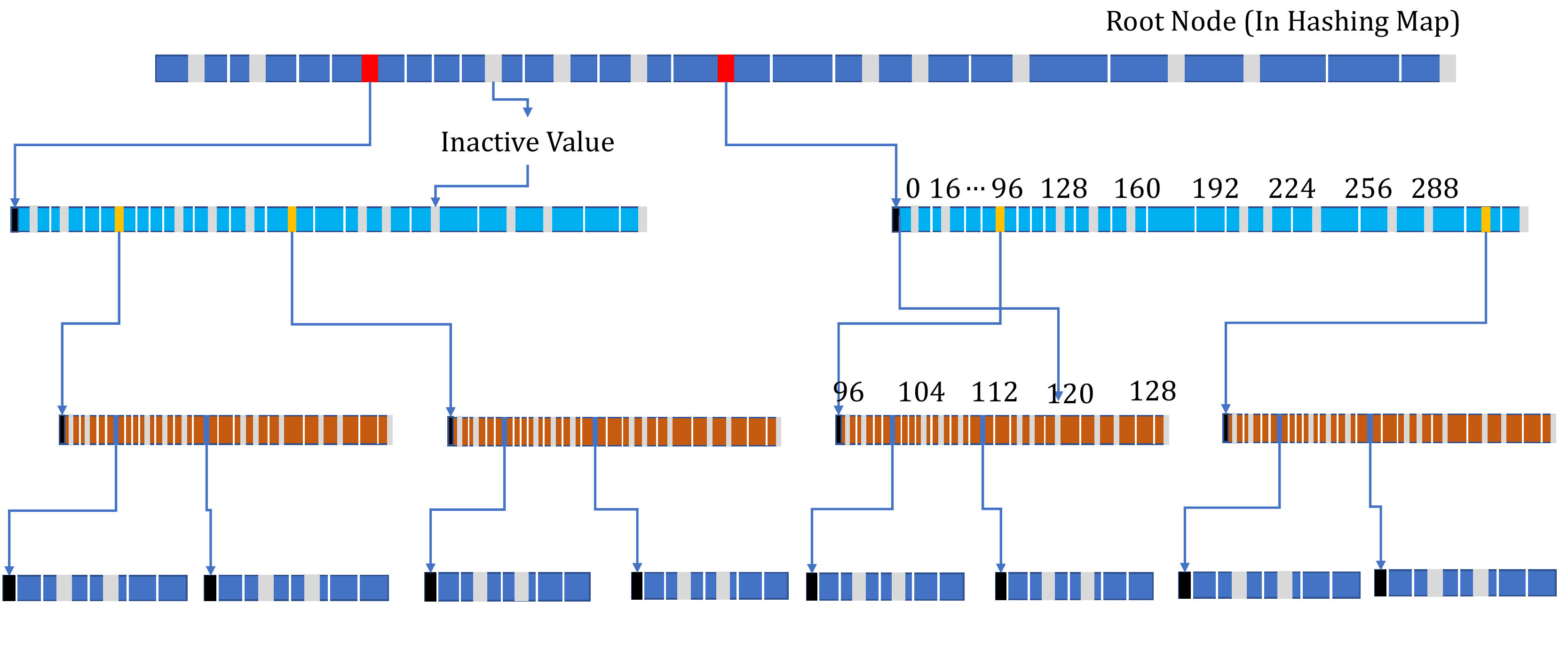}
\caption{The overall schematic of the adapted VDB-based data structure}
\label{vdb}
\vspace{-3mm}
\end{figure}

\begin{figure}[t]
\centering
\includegraphics[scale=0.388]{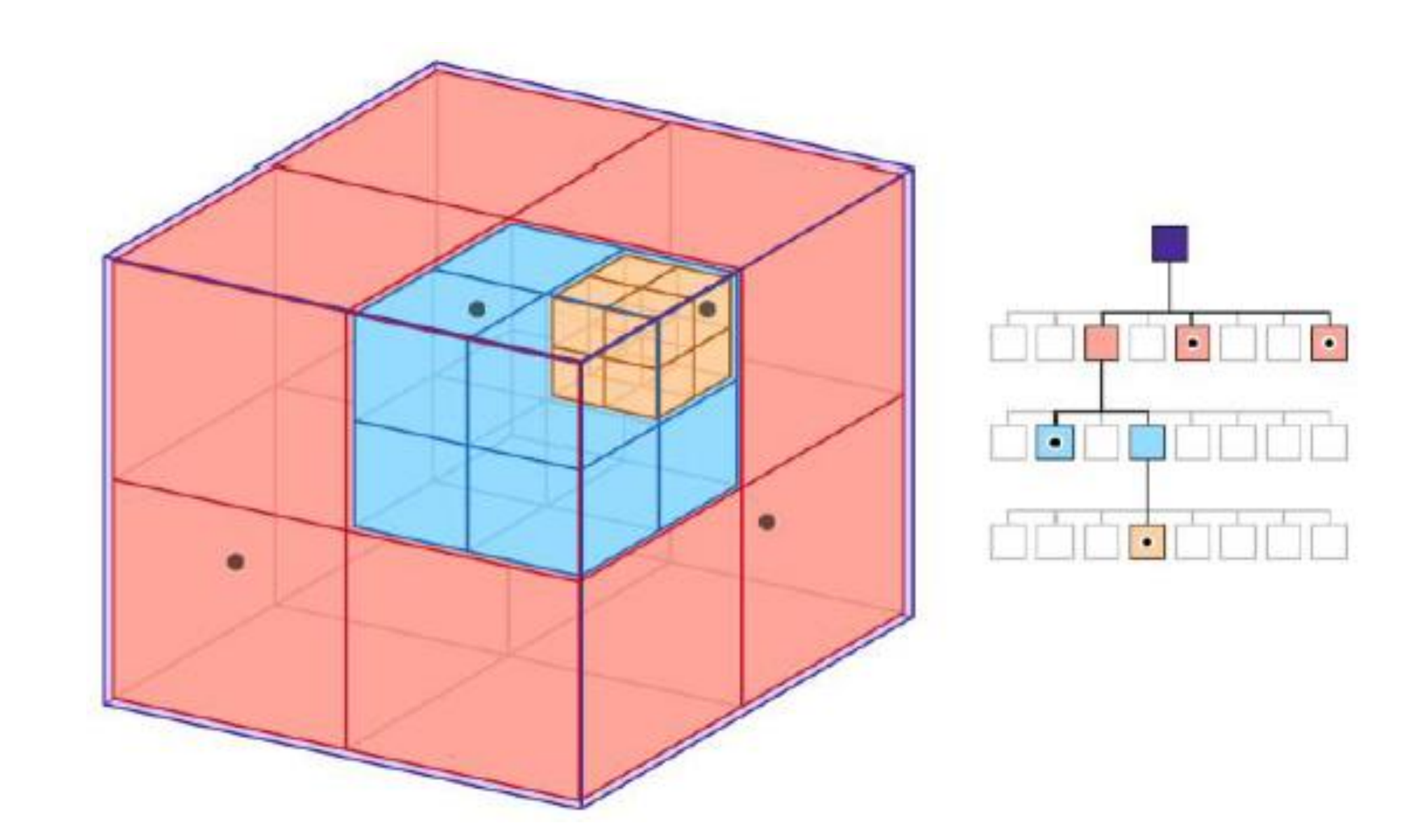}
\caption{The illustration of the voxelization for K-nearest neighbour query.}
\label{fig_knn}
\vspace{-1.5mm}
\end{figure}

\section{Proposed Methodology}
\subsection{The Overall System Framework}
\label{subsection_frame}
Recently, more advanced LiDAR SLAM systems have been proposed with higher efficiency and better performance. Our proposed framework is mainly based on typical ground vehicle LiDAR SLAM systems: LEGO-LOAM \cite{shan2018lego, liu2022light} and LIO-SAM \cite{shan2020lio}. 
Real-time localization can still be achieved because our proposed efficient data structure \textit{S-Voxel}. Furthermore, the proposed loop closure detection method greatly improves the overall accuracy of the entire LiDAR SLAM system.  We have proposed our own nearest neighbour query data strcuture termed \textit{S-voxel}, which has further enhance the speed of the the nearest neighbour query and finally the efficiency of the whole LiDAR-Inertial SLAM system.

\subsection{S-voxel Data Structure for Nearest Neighbour Search in LiDAR Point Clouds Registration}
\subsubsection{Motivations}
What can be further improved and investigated is mainly the nearest neighbor structure of LIO. The advantage of the KD-tree-like structure \cite{zhou2008real} is that the K-nearest neighbors can be queried strictly, which means the point to be queried is deterministic. The nearest neighbors can also be queried in the form of a range or box (range search/box search). In the query process, additional conditions such as the maximum distance are also set to achieve a fast approximate nearest neighbor (ANN) search. However, the traditional KD-tree has no incremental structure. Therefore, to improve the efficiency of the system, we need to redesign the algorithm and the searching methods. According to our experiments, the K-nearest neighbour is not truly required, and more efficient data structure should be designed.

\subsubsection{\textit{S-voxel} for voxel-based nearest neighbor}

In this work, we have proposed the sparse voxel based data structure to store the point clouds.  In practice point cloud registration often does not require strict K-nearest neighbors. If the K nearest neighbors find a very far point, it is unreasonable to use this point as a point-surface residual, and this part of the calculation is invalid. We might as well let the nearest neighbor structure itself have this search range restriction, and even if the KD-tree has a range restriction, it needs to be traversed node by node, which is obviously time-consuming. So we consider and propose a voxel-based nearest neighbor structure termed \textit{S-voxel}, which denote the voxel based data structure with sparsity. The structure of the proposed \textit{S-voxel} is shown in Fig. \ref{fig_svoxel}. The point clouds are stored in sparse voxels. The indices of the sparse voxels are put into an map without order. We have also compared our method with the VDB based data structure \cite{museth2013vdb}. The principles and the overall schematic of the VDB-based data structure are shown in Fig. \ref{vdb}. The VDB design is also very appropriate for the large-scale distance transformation problems, therefore we choose it for comparing of the near neibour query efficiency by the inference time of the whole SLAM system. The illustration of the voxelization K-nearest neighbour query is shown in Fig. \ref{fig_knn}. As shown in Fig. \ref{fig_knn}, the dividing of the space is based on the following principles: if the space contains points, we continue to divide the space; While if the space do not contain points, we then stop dividing the space.

Considering the sparseness of point clouds, we want voxels to also be stored in a sparse form. Voxels have some natural advantages: first, they naturally have the range limit of K nearest neighbors; second, they do not require additional operations during incremental construction, and it is convenient to delete them; third, the range of nearest neighbors can also be pre-defined, If you want to search more you can just do it. And if you want to accelerate, you can search less. Fourth, the voxel-based data structure is easy to be parallelized on GPU.

Therefore, we propose a sparse voxel-based nearest neighbor structure \textit{S-voxel}. We will find that this structure is more suitable for LIO, which can effectively reduce the time-consuming of point cloud registration, and will not affect the accuracy of LIO. We have proposed two versions of \textit{S-voxel}: One is linear and the other is based on space filling curves, which can be summarized as follows:

Our proposed \textit{S-voxel} is the composition of sparsely distributed voxels in space. There can be multiple points inside each voxel, and the grid coordinates of the voxel itself are mapped to the hash key value by the spatial hash function, and then a hash table is formed. Our proposed \textit{S-voxel} is composed of sparsely distributed voxels in space. There can be multiple points inside each voxel, and the grid coordinates of the voxel itself are mapped to the hash key value by the spatial hash function, and then a hash table is formed.

The hash function can be some certain formulations. Since we are actually storing 3D points, we can use a simple spatial hash. Denote the point clouds coordinate as $\textbf{h}=[h_1, h_2, h_3]^T$, and denote the size of the voxel as $l$. 
\begin{equation}
    \textbf{h} = [h_1, h_2, h_3]^T, \\ \textbf{v} = \frac{1}{s} [h_1, h_2, h_3]^T, 
\end{equation}
\begin{equation}
        id_{v}=hash(v)=(v_xn_x)\, xr \, (v_yn_y) \, xr \, (v_zn_z) \, mod \, N,
\end{equation}
where the $\textbf{h}$ is the 3D point, $v$ is the voxel mesh, $id$ is the Hashing key value.  $xr$ denotes the exclusive OR. The way the internal points in the voxels are stored is called its underlying structure. The simplest underlying structure is linear, which we call linear \textit{S-voxel}; if there are many points to be stored, we use spatial-filling curves to store them, which is called \textit{S-voxel-Curved}. Of course, in the LIO algorithm, we will avoid inserting a large number of points in the same voxel, which will affect the calculation efficiency. So the two algorithms are not very different in actual use. When searching for K-nearest-neighbors, we first calculate which specific voxel the searched point falls in, and then see if there is a valid \textit{S-voxel} within the predefined range. If so, take these points inside \textit{S-voxel} into account. We find at most K nearest neighbors within each \textit{S-voxel}, and then merge them. Linear \textit{S-voxel} only needs to traverse the interior points, then perform partial sorting, and then finally aggregate. 

\subsection{The K-nearest-neighbors Search}
When searching for the K nearest neighbors, we should first calculate which voxel the searched point falls in, and then see if there is a valid \textit{S-voxel} within the predefined range. If the valid \textit{S-Voxel} is found, we should take these points inside \textit{S-Voxel} into account. We find at most K nearest neighbors within each \textit{S-Voxel}, and then merge them. Linear \textit{S-Voxel} only needs to traverse the interior points, then perform partial sorting, and then aggregate. The polynomial case of \textit{S-Voxel}, termed \text{S-Voxel-Polynomial}, can find the nearest neighbor based on the index value on the space-filling curve, which is detailed as follows:

The \textit{S-Voxel-Polynomial} is a method to establish a mapping between high-dimensional data and low-dimensional data. Discrete \textit{S-Voxel-Polynomial} can be seen as the process of dividing the space into many small grids, and then finding an index for each grid. Therefore, when looking for the nearest neighbor of a point, you can first look at its index on the curve in this \textit{S-Voxel}, then find several neighbors around the index, and then return the result. When the number of points is large, the \text{S-Voxel-Polynomial} search will have less complexity than the linear search.

\subsection{The Incremental Map Updates}

The incremental map updates in the proposed are much simpler than kd-trees. In short, calculate the voxel grid corresponding to the increase point and add it directly. If it is the polynomial case, you also need to calculate the curve position of the polynomial case. Other than that, there is nothing else to do.

In Fast-LIO2 \cite{xu2021fast}, the system will delete a part of the historical point cloud and let the local map follow the vehicle forward. In the proposed \textit{S-voxel}, since traversing the entire \textit{S-voxel} local map is relatively slow, we make this process passive deletion instead of actively deleting after each frame calculation. So we add an LRU cache (least recently used) to the local map, and record which voxels have been used recently during the nearest neighbor search, so the less used voxels are naturally moved to the end of the queue. We will set the capacity of a local map, and when the maximum capacity is exceeded, those voxels that have not been used for a long time will be deleted. The delete operation is for the entire voxel, and the internal points will be deleted. This local map caching strategy will also allow the local map to follow the vehicle movement with fewer places to actually operate.

\begin{figure}[t]
\centering
\includegraphics[scale=0.156]{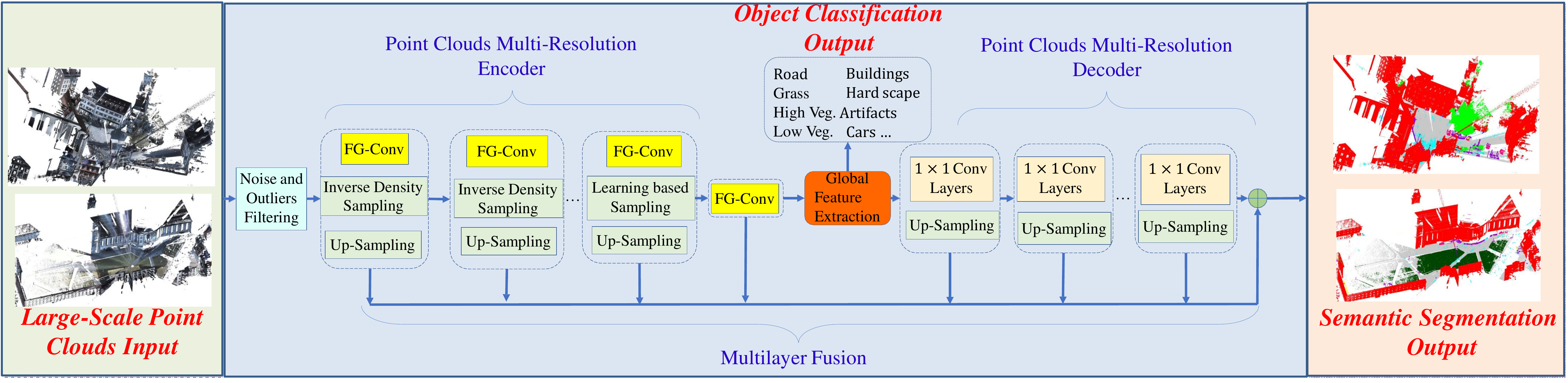}
\caption{The whole network architecture of our proposed FG-Net \cite{liu2022fg} network framework, which main constituting components is the FG-Conv \cite{liu2021fg} network operators. The network framework can be utilzed to do loop closure detection effectively and efficiently.} 
\label{fig_fg_framework}
\vspace{-7mm}
\end{figure}


\begin{table}[bp!]
\caption{The Comparisons of Localization Accuracy and Computational Cost tested on Hong Kong Science Park shown in Fig. 1 in the Range of around 1000m $\times$ 150m. We Have Compared the Efficiency of the Proposed \textit{S-Voxel} and VDB-based Data Structure in Efficiency.}
\label{table_full}
\begin{center}
\scalebox{0.92}{\begin{tabular}{ccc}
\toprule

Methods & Computational Time (ms/frame) & Error (cm)\\
\hline
Ours (w/o \textit{LC-Net})&57.62&13.45\\
Ours (w/o \textit{S-Voxel})&102.58&12.21\\
Ours (Full)&62.36&11.96\\
LEGO-LOAM \cite{shan2018lego}&97.9&13.92\\ 
LIO-SAM \cite{shan2020lio}&92.3&13.13\\
Fast-LIO2 \cite{xu2021fast}&91.62&13.07\\
\bottomrule
\end{tabular}}
\end{center}
\vspace{-3mm}
\end{table}

\begin{table}[bp!]
\caption{The Comparisons of Localization Accuracy and Computational Cost with Various of Data Structures Tested on Hong Kong Science Park Shown in Fig. 1 in the Range of around 1000m $\times$ 150m. We have Compared the Efficiency of the Proposed \textit{S-Voxel} and VDB-Based Data Structure in Efficiency.}
\label{table_data}
\begin{center}
\scalebox{0.9}{\begin{tabular}{ccc}
\toprule

Methods & Computational Time (ms/frame) & Error (cm)\\
\hline
Ours (\textit{VDB-based} \cite{museth2013vdb})&88.58&12.25\\
Ours (\textit{KD-Tree-based} \cite{chen2019fast})&91.62&12.31\\
Ours (\textit{S-Voxel}-based) (Full)&62.36&11.96\\
\bottomrule
\end{tabular}}
\end{center}
\vspace{-3mm}
\end{table}


\begin{table}[bp!]
\caption{The Comparisons Results of the Loop Closure Test Success Rate at the Tunnel Environments}
\label{table_loop_close}
\begin{center}
\scalebox{1.00}{\begin{tabular}{ccc}
\toprule

Methods & Success & Failure in Loop Closure\\
\hline
Ours (w/o \textit{LC-Net})&5&2\\
Ours (w/o \textit{S-Voxel})&7&0\\
Ours (Full)&7&0\\
LPD-net \cite{liu2019lpd} &5&2\\
LEGO-LOAM \cite{shan2018lego}&2&5\\ 
LIO-SAM \cite{shan2020lio}&3&4\\
Fast-LIO2 \cite{xu2021fast} & 4 & 3 \\

\bottomrule
\end{tabular}}
\end{center}
\vspace{-3mm}
\end{table}


\begin{figure*}[ht]
\centering
\includegraphics[scale=0.51]{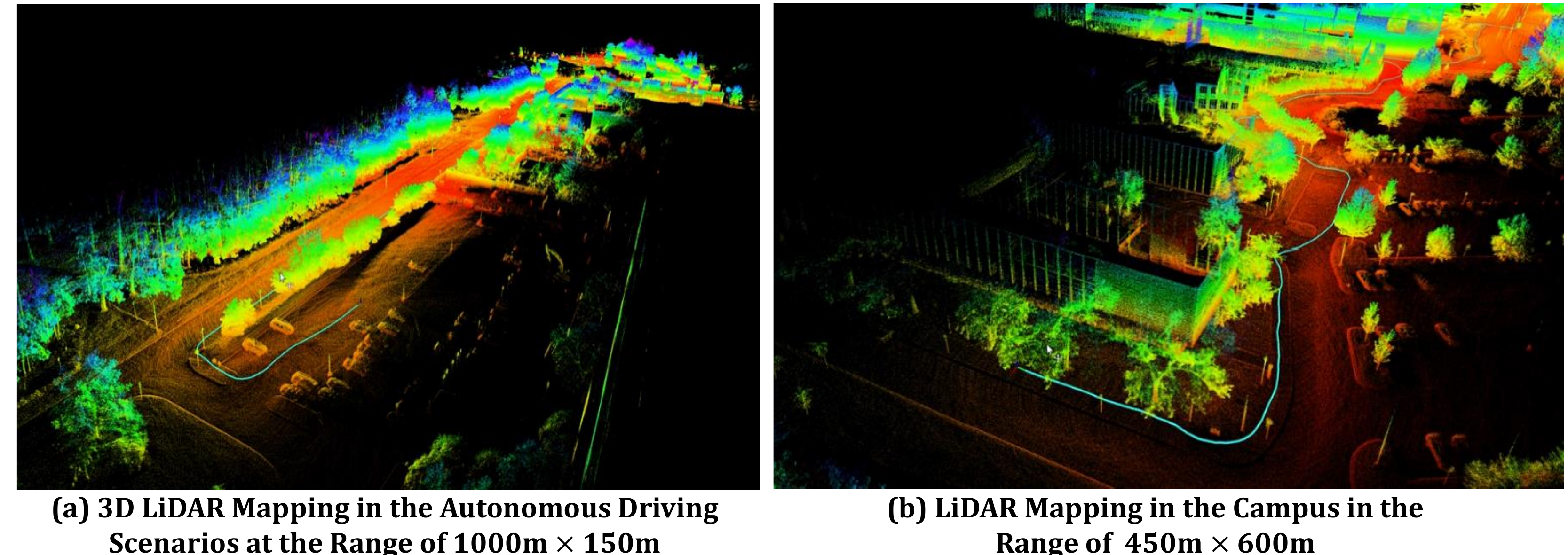}
\caption{The final results of the LiDAR-Inertial mapping results in various of real-site complicated environment. Subfigure (a) shows the mapping results of our proposed LiDAR-inertial SLAM system at the autonomous driving scenarios at the range of $1000m \times 150m$. Subfigure (b) shows the 3D mapping results of our proposed LiDAR-inertial SLAM.}
\label{fig_tunnel}
\vspace{-6mm}
\end{figure*}
    
\subsection{The Learning based Loop Closure}
For the learning-based loop closure, we have utilized our proposed effective FG-Net \cite{liu2022fg, liu2022weaklabel3d, liu2022weakly}  for the task of recognizing candidates for loop closure. The detailed network architecture is shown in Fig. \ref{fig_fg_framework}. After the candidate scans for loop closure are found, the optimization can be performed to correct the large-scale drifts in mapping of the environment.

\section{Experiments}

\subsection{Qualitative Experimental Results}
 In our setting of the training of the network of the loop closing network \textit{FG-LC-net}, we have adopted a learning rate of $5 \times 10^{-4}$ with the weight decay of 0.985 in each training epoch. And the training of our proposed network \textit{FG-LC-net} lasts for 280 epoches. We implement the network framework using \textit{Pytorch}. The final real-site mapping results are shown in Fig. \ref{fig_tunnel}. Subfigure (a) shows the mapping results of our proposed LiDAR-inertial SLAM system at the autonomous driving scenarios at the range of $1000m \times 150m$. Subfigure (b) shows the 3D mapping results of our proposed LiDAR-inertial SLAM. Subfigure (c) shows the real-site mapping results of our Proposed LiDAR-inertial SLAM system with fused color from the camera. Subfigure (d) shows the real-site mapping results during the experiments at the Hong Kong local tunnel environments. It can be demonstrated that our proposed LiDAR-Inertial mapping system can perform large-scale mapping at various complicated circumstances. The mapping results at the autonomous driving scenarios show that the proposed LiDAR-Inertial SLAM system can provide accurate mapping results for the objects with fine-grained details, such as the roadside trees. Also, we can see the driving road very explicitly in the campus environment. Also, for the applications of the tunnel inspections, we can provide very detailed reconstruction of the walls inside tunnel. Therefore, it can be seen qualitatively that the proposed LiDAR-Inertial SLAM system has high accuracy and great robustness under different large-scale complicated test circumstances.

 
 \subsection{Quantitative Experimental Results}
 As shown in Table \ref{table_full}, our proposed neighbour query strategy \textit{S-Voxel} has great boost on the efficiency of the whole SLAM system. And our proposed loop closure detection network \textit{FG-LC-Net} can significantly enhance the global localization accuracy of the whole SLAM system with accurately detection of the loop closure. Compared with previous SOTAs LIO-SAM \cite{shan2020lio} and LEGO-LOAM \cite{shan2018lego}, our proposed method has out-performed them in terms of both accuracy and efficiency. We have also done ablation studies for the optimization function components of the proposed  \textit{FG-LC-Net}. 
  As shown in Table \ref{table_data}, our proposed \textit{S-Voxel} is better comapred with the counter part VDB \cite{museth2013vdb} and KD-Tree \cite{chen2019fast}. As shown in Table \ref{table_loop_close}, with our proposed learning based loop closure detection approach, the success rate of the loop closure is significantly improved. In addition, the robustness of the whole SLAM system is guaranteed. Also, our proposed method has outperformed the previous SOTAs LPD-Net \cite{liu2019lpd}. It has also been demonstrated in Table \ref{table_full} that our proposed \textit{LC-Net} only add marginal computational cost based on the LiDAR SLAM system, which means high efficiency and real-time performance of the proposed LiDAR-Inertial SLAM system can still be realized.

\subsection{Real-Site Robot Navigation Integrated LiDAR SLAM System}
    Finally, The constructed 3D map by our proposed LiDAR-Inertial SLAM system is also of great significance to the further potential applications such as large-scale 3D building reconstructions, 3D model analysis, and also the building defects localization and detection. We have also deployed our system into the project of tunnel inspections, the related results are shown in Fig. \ref{fig_tunnel}.  As shown in Fig. \ref{fig_tunnel}, we have also deployed our system to the local autonomous surveillance and inspections at Hong Kong. Our proposed LiDAR-Inertial SLAM system can be integrated seamlessly with the motion planning approaches to realize fully autonomous navigation for both the autonomous unmanned aerial vehicle and the autonomous unmanned ground vehicle in the narrow corridors and complex tunnel environments.    
\section{Conclusions}
In this work, we have proposed a LiDAR-Inertial fusion-based simutaneous localization and mapping system. We have proposed a data structure termed \textit{S-Voxel} to accelerate the nearest neighbour query process.  It has been demonstrated that the real-time performance of the LiDAR-Inertial SLAM system can be greatly improved. And by the proposed learning based loop closure detection approaches, we can further enhance the accuracy of the final global mapping results. It can be demonstrated that our proposed LiDAR-Inertial SLAM system show great performance and robustness in both indoor and outdoor circumstances in diverse complicated circumstances including the real road scenarios and the underground tunnel circumstances.

\addtolength{\textheight}{0cm}   





\bibliographystyle{IEEEtran}
\bibliography{references}

\end{document}